\documentclass[runningheads]{llncs}

\usepackage{graphicx}
\usepackage[utf8]{inputenc}
\usepackage[english]{babel}
\usepackage{longtable,booktabs}
\usepackage{xcolor,colortbl}
\usepackage{multirow}
\usepackage{tabularx}
\usepackage{blindtext}
\usepackage{amsmath}

\usepackage{soul}

\usepackage{subcaption}

\renewcommand{\vec}{\mathbf}

\begin{document}
\title{
    AT-ST: Self-Training Adaptation Strategy for OCR in Domains with Limited Transcriptions
}
\titlerunning{AT-ST: Self-Training Adaptation Strategy for OCR}
%
\author{Martin Kišš\orcidID{0000-0001-6853-0508} \and
Karel Beneš\orcidID{0000-0002-0805-1860} \and \\
Michal Hradiš\orcidID{0000-0002-6364-129X}}
%
\authorrunning{M. Kišš et al.}
%
\institute{Brno University of Technology, Czechia \\
\email{\{ikiss,ibenes,hradis\}@fit.vutbr.cz}}
%
\maketitle              
\begin{abstract}

This paper addresses text recognition for domains with limited manual annotations by a simple self-training strategy.
Our approach should reduce human annotation effort when target domain data is plentiful, such as when transcribing a collection of single person's correspondence or a large manuscript.
We propose to train a seed system on large scale data from related domains mixed with available annotated data from the target domain.
The seed system transcribes the unannotated data from the target domain which is then used to train a better system.
We study several confidence measures and eventually decide to use the posterior probability of a transcription for data selection.
Additionally, we propose to augment the data using an aggressive masking scheme.
By self-training, we achieve up to 55\,\% reduction in character error rate for handwritten data and up to 38\,\% on printed data.
The masking augmentation itself reduces the error rate by about 10\,\% and its effect is better pronounced in case of difficult handwritten data.

\end{abstract}

\vspace{-2.5em}
\begin{figure}[h!]
    \centering
    \includegraphics[width=0.84\linewidth]{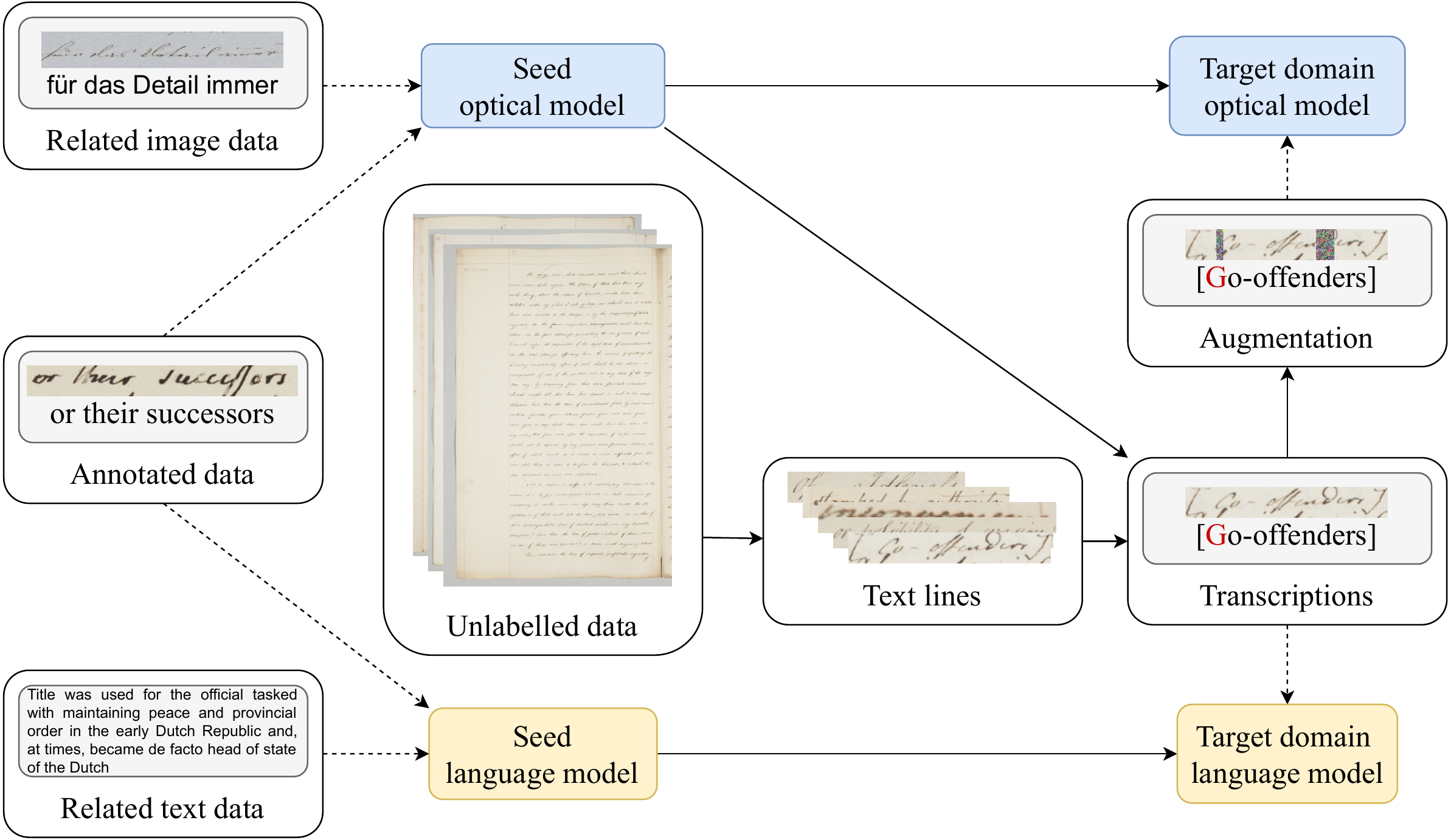}
    \caption{%
        Proposed OCR system adaptation pipeline to a target domain.
    }
    \label{fig:pipeline}
    \vspace{-2.8em}
\end{figure}

\section{Introduction}

When transcribing documents from a specific \emph{target domain}, e.g.\ correspondence of a group of people or historical issues of some periodical, the accuracy of pre-trained optical character recognition (OCR) systems is often insufficient and the automatic transcriptions need to be manually corrected.
The manual effort should naturally be minimized.
One possibility is to improve the OCR system through adapting it using the manually corrected transcriptions.
In this paper, we propose to improve the OCR system more efficiently by utilizing the whole target domain, not only the manually corrected part.


We explore the scenario where a small part of the target domain is manually transcribed (annotated) and a large annotated dataset from \emph{related domains} is available (e.g. large-scale public datasets, such as READ~\cite{sanchez_icdar2017_2017} or IMPACT~\cite{papadopoulos_impact_2013}).
Our pipeline starts by training a \emph{seed} system on this data.
The seed system transcribes all unannotated text lines in the target domain, and the most confident transcriptions are then used as a ground truth to retrain and adapt the system.


While our approach does not depend on the actual implementation of the OCR system and should be applicable to a broad range of methods, we limit our experimentation to a combination of a CTC-based~\cite{GravesCTC} optical model with an explicit language model.
We show that:
\begin{enumerate}
    \item The posterior probability of a transcription produced by prefix search decoder on top of a CTC optical model is a well-behaved OCR confidence measure that allows to select lines with low error rate.
    \item A masking data augmentation scheme helps the optical model to utilize the machine-annotated data.
    \item Even when well pre-trained on a related domain, both the optical and the language model benefit significantly from machine annotated data from the target domain and the improvements are consistent over a range of overall error-rates, languages, and scripts.
\end{enumerate}

\section{Related work}

Semi-supervised learning approaches in OCR are based on \emph{Pseudo-Labeling}~\cite{nagai_recognizing_2020} or \emph{Self-Training}~\cite{stuner_self-training_2017,das_adapting_2020}.
They differ in that pseudo-labeling continues the training with the seed system, whereas in self-training, a new model is trained on the extended data~\cite{xie_self-training_2020}.
However, these terms are often mixed at authors' liberty.

Automatic transcriptions naturally contain errors, but consistent data is desirable for training recognition models.
Thus, the goal is to either correct these errors or select only text lines that do not contain these errors.
If multiple models are trained on annotated data, it is possible to combine the outputs of these models to obtain a more reliable transcript~\cite{nagai_recognizing_2020,frinken_bunke}.
If there is only one trained model, transcriptions can be checked against a language model~\cite{leifert_two_2020}, a lexicon~\cite{stuner_self-training_2017}, or page-level annotations~\cite{leifert_two_2020} if available.

Semi-supervised training of language models is not explored very well in literature.
It has been shown that small gains can indeed be achieved with count based n-gram LMs trained on transcripts produced by a speech recognition system~\cite{egorova-spanish-ss-lm}.
Machine annotations are a bit more common in discriminative language modeling, where the LM is trained to distinguish a hypothesis with low error rate among others.
This work typically focuses on getting a diverse enough training data for the training~\cite{celebi-ss-disc-lm}.
Most recent LM literature focuses on training universal models on vast amounts of data, with the hope that such LM can adapt to anything~\cite{brown-gpt3}.

In a recent paper, an approach similar to ours was proposed~\cite{das_adapting_2020}.
The authors propose an iterative self-training scheme in combination with extensive fine-tuning to improve a CTC-based OCR.
However, their experiments are limited to printed text, they recognize individual words and they achieve negligible improvements from the self-training itself.
Also, no attention is paid to adaptation of language models.

Data augmentation is a traditional approach to reduce overfitting of neural networks.
In computer vision, the following are commonly used~\cite{dutta_improving_2018,wigington_data_2017,das_adapting_2020}: affine transformations, change of colors, geometric distortions, adding noise, etc.
A technique similar to our proposed masking is \emph{cutout}~\cite{devries_improved_2017}, which replaces content of random rectangle areas in the image by gray color.
To the best of our knowledge, it has never been used for text recognition.

\section{AT-ST: Adapting Transcription system by Self-Training}
\label{sec:semisup-learning}
Our goal is to obtain a well performing OCR system for the \emph{taget domain} with little human effort and when there is the following available:
(1) \emph{Related domain data}, which loosely matches the style, language or overall condition of the target domain; (2) \emph{annotated target domain data}, which has human annotations available.
Together, these two provide the \emph{seed data}.
The rest of the target domain constitutes (3) \emph{unannotated target domain data}, which we want to utilize without additional human input.
Eventually, we measure the performance of our systems by Character Error Rate (CER) on a held out \emph{validation} and \emph{test} portion of the annotated target domain data.

As summarized in Figure~\ref{fig:pipeline}, our pipeline progresses in four steps:
(1) A seed OCR system is trained on the seed data.
(2) This OCR is used to process all unannotated lines from the target domain.
(3) For each of these lines, confidence score is computed using a suitable OCR confidence measure.
Then, a best-scoring portion of a defined size is taken as the \emph{machine annotated} (MA) data.
(4) MA data is then merged with the seed data and a new OCR is trained on it.
In general, steps 2\,--\,4 can be repeated, which might lead to further adaptation of the OCR to the target domain.

\subsection{Implementing OCR system}\label{sec:ocr-system-design}
Our pipeline could in principle be implemented with any type of optical and language model.
In our implementation, the optical model is a neural network based on the CRNN architecture~\cite{shi_end--end_2017} trained with the CTC loss function~\cite{GravesCTC}.
This means that the optical model transforms a 2-D image of a text line $\ell$ into a~series of \emph{frames} $\vec{f}_1, \ldots, \vec{f}_T$, where every frame $\vec{f}_t$ is a vector of probabilities over an alphabet including a special \emph{blank} symbol representing empty output from the frame.

It is viable to obtain the transcription by simply taking the most probable character from each frame (greedy decoding).
More accurate results can be obtained by \emph{prefix search decoding}~\cite{GravesCTC}, which accommodates the fact that the probability of a character may be spread over several frames.
The prefix search decoding keeps a pool of the most probable partial transcripts (prefixes) and updates their probabilities frame by frame.
All possible single-character prefix extensions are considered at every frame and only the most probable prefixes are retained.

This approach can be readily extended by introduction of a language model that estimates the probability of a sequence of characters.
The total score of prefix $a$ is then computed as:
\begin{equation}\label{eq:decoder}
    \log S(a) = \log S_O(a) + \alpha \log S_L(a) + \beta |a|
\end{equation}

where $S_O(a)$ is the score of $a$ as given by the optical model, $S_L(a)$ is the language model probability of $a$, $\alpha$ is the weight of the language model, $|a|$ is the length of $a$ in characters, and $\beta$ is the empirically tuned insertion bonus.
Since we use an autoregressive language model that estimates probability of every character given the previous ones, we introduce the additional terms on-the-fly during the search.
A total of $K$ best scoring prefixes is kept during decoding.

\subsection{OCR confidence measures}\label{sec:confidences-def}
To identify the most accurately transcribed lines, we propose several confidence measures $M$.
The output of a measure $M$ is a score in range [0, 1].
The score serves as the predictor of error rate of the greedy hypothesis $h_g$.

\paragraph{CTC loss} is the probability $P(h_g|\ell)$ as defined by the CTC training criterion, i.e. the posterior probability of the hypothesis $h_g$ marginalized over all possible alignments of its letters to the frames. 
We normalize this probability by the length of the hypothesis in characters.

\paragraph{Transcription posterior probability} is estimated using prefix decoding algorithm using only the optical scores, i.e. with $\alpha = \beta = 0$.
For each\footnote{Except lines with only one frame, where there may be fewer prefixes considered.} line $\ell$, the decoding produces a set of hypotheses $\mathcal{H} = \{h_1, \ldots, h_K\}$ together with their associated logarithmic unnormalized probabilities $C(h_i)$.
We normalize these scores to obtain posterior probabilities of those hypotheses:
\begin{equation}\label{eq:postrior-proba}
    M_{\mathrm{posterior}}(\ell) = P(h_g|\ell) = \frac{exp(C(h_g))}{\sum_{i=1} exp(C(h_i))}
\end{equation}
We assign $P(h | \ell) = 0$ to hypotheses outside of $\mathcal{H}$.

Additionally, we propose four measures that focus on the maximal probabilities in each frame.
This should reflect the fact that greedy decoding -- the source of the hypothesis -- also begins by taking these maxima.
Denoting these maxima as $m_t$, we define:

\begin{itemize}
    \item \emph{probs mean} as the average of $m_t$ for all $t \in \{1, \ldots, T\}$.
    
    \item \emph{char probs mean} as the average of $m_t$ for those $t$ where blank is not the most probable output.
    
    \item \emph{Inliers Rate} is based on the distribution of the maxima from an examined data set.
    This is modelled by a maximum likelihood Gaussian.
    Then, the confidence score of each text line is computed as the ratio of $t$ such, that $m_t$ lies within $2\sigma$.
    
    \item \emph{Worst best} finds the Viterbi alignment of the hypothesis to the output probabilities.
    Then for each character in the hypothesis, it takes a corresponding frame with the highest $m_t$.
    Finally, the confidence score is the minimum of the values.
\end{itemize}

\subsection{Masking augmentation}
\begin{figure}[t]
    \centering
    \includegraphics[height=12pt]{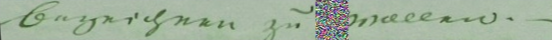} \hfill
    \includegraphics[height=12pt]{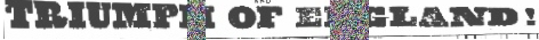} \\
    \includegraphics[height=12pt]{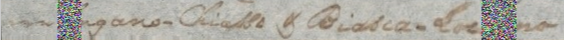} \hfill
    \includegraphics[height=12pt]{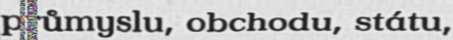} \\
    \includegraphics[height=12pt]{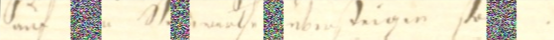} \hfill
    \includegraphics[height=12pt]{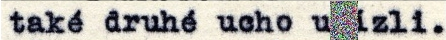} \\
    \includegraphics[height=12pt]{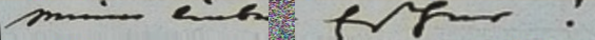} \hfill
    \includegraphics[height=12pt]{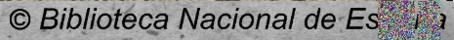}
    \caption{Example of the proposed masking augmentation}
    \label{fig:masking}
\end{figure}

To provide a more challenging learning environment, we propose to mask parts of the input image out.
Examples are shown in Figure \ref{fig:masking}.
Since the transcription of the image remains unchanged, we believe this technique might especially improve the implicit language model inside the optical network.

The masking augmentation replaces random parts of the input image by noise with uniform distribution.
The number of masked regions is sampled from binomial distribution.
The masked regions span full height of a text line and their width is sampled from another uniform distribution.

We chose the random noise, instead of for example constant color, to clearly mark the masked regions. 
We believe that conveying this information to the model should avoid ambiguity between empty and masked regions, reduce the complexity of the training task, and consequently improve convergence and the final accuracy. 

\section{Datasets}
\begin{figure}[t]
    \centering
    \subcaptionbox{%
        READ
        \label{fig:line-example-read}}
        {\includegraphics[width=0.48\linewidth]{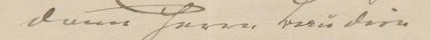}
    }
    \hfill
    \subcaptionbox{%
        IMPACT
        \label{fig:line-example-impact}}
        {\includegraphics[width=0.48\linewidth]{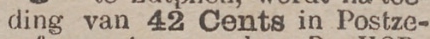}
    }
    \\[0.5em]
    \subcaptionbox{%
        Bentham (annotated)
        \label{fig:line-example-bentham-annotated}}
        {\includegraphics[width=0.48\linewidth]{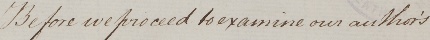}
    }
    \hfill
    \subcaptionbox{%
        Czech news (annotated)
        \label{fig:line-example-ln-annotated}}
        {\includegraphics[width=0.48\linewidth]{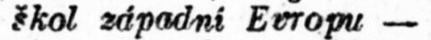}
    }
    \\[0.5em]
    \subcaptionbox{%
        Bentham (unannotated)
        \label{fig:line-example-bentham-unannotated}}
        {\includegraphics[width=0.48\linewidth]{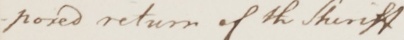}
    }
    \hfill
    \subcaptionbox{%
        Czech newspapers (unannotated)
        \label{fig:line-example-ln-unannotated}}
        {\includegraphics[width=0.48\linewidth]{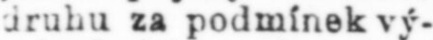}
    }
    \\
    \caption{%
        Example text lines by datasets.
    }
    \label{fig:line-examples}
\end{figure}

We performed experiments on printed and handwritten data.
The description of individual datasets is in the following paragraphs and the sizes of the datasets are summarized in Table~\ref{tab:datasets} and text line examples from each dataset are depicted in Figure~\ref{fig:line-examples}.

\begin{table}[t]
    \centering
    \caption{%
        Size of datasets in lines.
    }\label{tab:datasets}
    \begin{tabular}{
        @{\extracolsep{6pt}}@{\kern\tabcolsep}
        lrrrrrr}
        \toprule
                    & \multicolumn{3}{c}{Handwritten datasets} & \multicolumn{3}{c}{Printed datasets} \\
                        \cline{2-4}
                        \cline{5-7}
                        \addlinespace[0.1cm]
                    & \multicolumn{1}{c}{Training}
                    & \multicolumn{1}{c}{Validation}
                    & \multicolumn{1}{c}{Test}
                    & \multicolumn{1}{c}{Training}
                    & \multicolumn{1}{c}{Validation}
                    & \multicolumn{1}{c}{Test} \\
        \midrule
        Related domain        &    189\,805 &   \multicolumn{1}{c}{--}   & \multicolumn{1}{c}{700} & 1\,280\,000 &   \multicolumn{1}{c}{--}   & 12\,000 \\
        Target d. annotated   &      9\,198 & \multicolumn{1}{c}{1\,415} & \multicolumn{1}{c}{860} &     32\,860 & \multicolumn{1}{c}{6\,546} &  1\,048 \\
        Target d. unannotated & 1\,141\,566 &   \multicolumn{1}{c}{--}   &  \multicolumn{1}{c}{--} & 2\,673\,626 &   \multicolumn{1}{c}{--}   &    \multicolumn{1}{c}{--}   \\
        \bottomrule
    \end{tabular}
\end{table}

As we used several datasets in the experiments, we needed to ensure consistency in terms of baseline positioning and text line heights.
We used text line detection model to detect text lines in all datasets and mapped the produced text lines to the original transcriptions based on their content and location if available.
The detection model is based on ParseNet architecture \cite{kodym_page_2021} and was trained on printed and handwritten documents from various sources.

\paragraph{Handwritten datasets}
For experiments with handwritten data, we used the ICDAR 2017 READ Dataset \cite{sanchez_icdar2017_2017}, the ICFHR 2014 Bentham Dataset \cite{sanchez_icfhr2014_2014}, and additional unannotated pages (\emph{Unannotated Bentham Dataset}) obtained from the Bentham Project\footnote{\texttt{https://www.ucl.ac.uk/bentham-project}} as the related domain, target domain annotated and target domain unannotated datasets respectively.
The READ Dataset contains pages written in German, Italian, and French, while the Bentham Dataset and the Unannotated Bentham Dataset consist of pages written in early 19th-century English.
In contrast to the other datasets with plentiful writer, majority of the dataset has been written by J.~Bentham himself or his secretarial staff.

\paragraph{Printed datasets}
For experiments with printed data, we used IMPACT Da\-ta\-set~\cite{papadopoulos_impact_2013} as the related domain data.
As target domain, we used historical printings of Czech newspapers\footnote{\texttt{https://www.digitalniknihovna.cz/mzk}}.
This data is generally of relatively low quality as it was scanned from microfilms.
Approximately 2000 pages of it were partially transcribed by volunteers, these lines serve as the annotated target domain data.
In addition, we checked the test set and removed text lines with ambiguous transcriptions.
The rest of the pages was used as the target domain unannotated data.
From the IMPACT dataset, we randomly sampled 1.3M text lines.
The most common languages represented in the IMPACT dataset are Spanish, English, and Dutch, while the Czech newspapers dataset contains pages written in Czech.

\subsection*{Related domain data for language models}
It is reasonable to expect availability of contemporary text data for any language with writing system.
To reflect that in our experiments, we provide pre-training corpora to our language models.
This way, the overall structure of data for the optical model and the language model is the same.
We do not attempt to use the training transcriptions from the related domain OCR data, as those come in various languages and are orders of magnitude smaller than available text data. 

For English, we use the raw version of Wikitext-103~\cite{MerityPointerSentinel}, which contains ca. 530M characters (103 M words).
Since the dataset is originally prepared for word-level language modeling based on whitespace-delimited tokenization, we adjusted the spacing around punctuation to follow the standard rules.

For Czech, we use an in-house corpus consisting mainly of news from internet publishing houses and wikipedia articles.
The total size is approx. 2.2\,B characters (320\,M words).

\section{Experiments}
Experimentally, we demonstrate the effectiveness of both the proposed self-training scheme and the masking augmentation.
The first step of our self-training pipeline is training seed optical models on human annotated data.
Using their outputs, we explore the predictive power of the proposed confidence measures and estimate the error rate of the untranscribed data in Section~\ref{sec:confidences-exp}.
Then in Section~\ref{sec:ss-exp}, we show the effect of adding this machine annotated data into the training process of both the optical model and the language model.
As all the above mentioned optical models are trained with our proposed masking augmentation, we finally explore its effect separately in Section~\ref{sec:masking-exp}.

To simulate the effect of investing more or less effort into manually transcribing the target domain, we consider two conditions for each dataset:
\emph{Big}, which is equal to using all of the annotated target domain data, and \emph{small}, which is simulated by randomly taking 10\,\% of it\footnote{Specifically, we do the subsampling on the level of pages, to emulate the possible effect of reduced variability in data.}.
When training the optical model in the big setup, we give extra weight to the text lines from the target domain so that the model adapts to it more.
As a result, we have four different setups of seed data --- \emph{handwritten small}, \emph{handwritten big}, \emph{printed small}, and \emph{printed big}.

\subsection{Specification of optical models and language models}

The optical model is a neural network based on the CRNN architecture~\cite{shi_end--end_2017}, where convolutional blocks are followed by recurrent layers and the network is optimized using the CTC loss~\cite{GravesCTC}.
To speed up the training process, we initialize the convolutional layers from a pretrained VGG16\footnote{From PyTorch module \texttt{torchvision.models.vgg16}}.
The recurrent part consists of six parallel BLSTM layers processing differently downsampled representations of the input.
Their outputs are upsampled back to the original time resolution, summed up, and processed by an additional BLSTM layer to produce the final output.
In all experiments, the input of the optical model is an image $W \times 40$, where $W$ is its width and the height is fixed to 40 pixels, and the output is matrix of dimensions $W/4 \times |V|$, where $|V|$ is the size of the character vocabulary including the blank symbol.
The exact definition of the architecture is public\footnote{\texttt{https://github.com/DCGM/pero-ocr, }}.

Each optical model was optimized using Adam optimizer for 250k iterations. The initial learning rate was set to $2\times10^{-4}$ and was halved after 150k and 200k iterations.
Except for the experiments in Section~\ref{sec:masking-exp}, the masking augmentation was configured as follows:
The number of masked regions was sampled from a binomial distribution with the number of trials equal to the width of the input image in pixels, and the success probability equal to $5\times 10^{-3}$.
The width of the masked region was sampled uniformly from interval [5, 40], making the largest possible region a square.

All language models (LMs) were implemented as LSTMs~\cite{Sundermeyer2012LSTM}, with 2 layers of 1500 units.
Input characters are encoded into embeddings of length 50.
We optimize the LMs using plain SGD without momentum, with initial learning rate 2.0, halving it when validation perplexity does not improve.
For validation, we always use the reference transcriptions of the respective validation data.
We train all our LMs using BrnoLM toolkit\footnote{\texttt{https://github.com/BUTSpeechFIT/BrnoLM}}.

When doing the prefix decoding, we tune the LM weight $\alpha$ in range 0.0\,--\,1.5 and the maximal number of active prefixes $K$ up to 16.
To keep the hyperparameter search feasible, the character insertion bonus $\beta$ is kept at 1.0.
This value is a~result of preliminary experiments: lower values lead invariably to increased number of deletions, whereas higher values resulted in a trade-off between insertions and deletions, and a slow but steady increase in total error rate.

\subsection{Predictive power of confidence measures}\label{sec:confidences-exp}

\begin{figure}[p]
    \centering
    \subcaptionbox{%
        Handwritten, small\label{fig:conf-bentham-small}}
        {\includegraphics[width=0.45\linewidth]{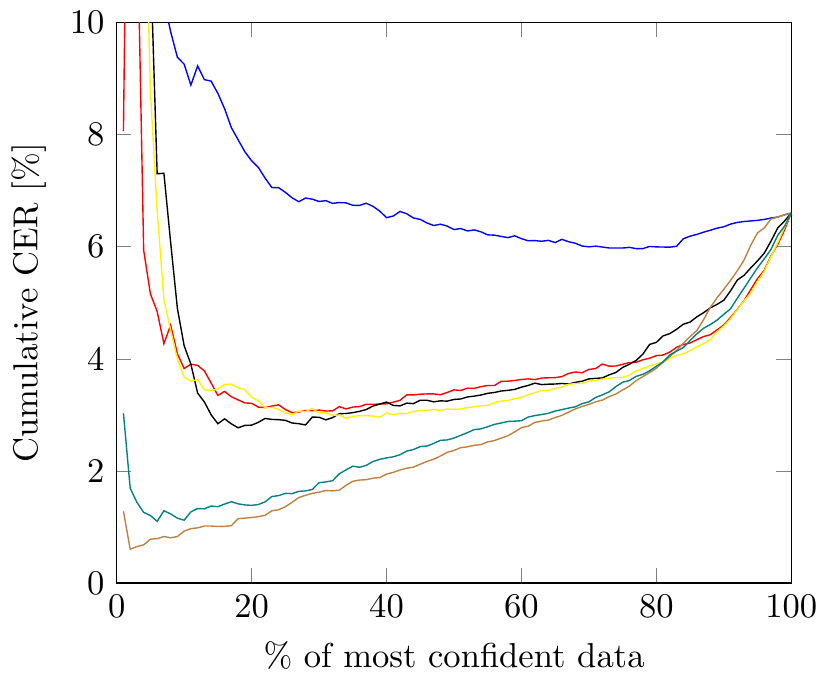}
    }
    \qquad
    \subcaptionbox{%
        Handwritten, big\label{fig:conf-bentham-big}}
        {\includegraphics[width=0.45\linewidth]{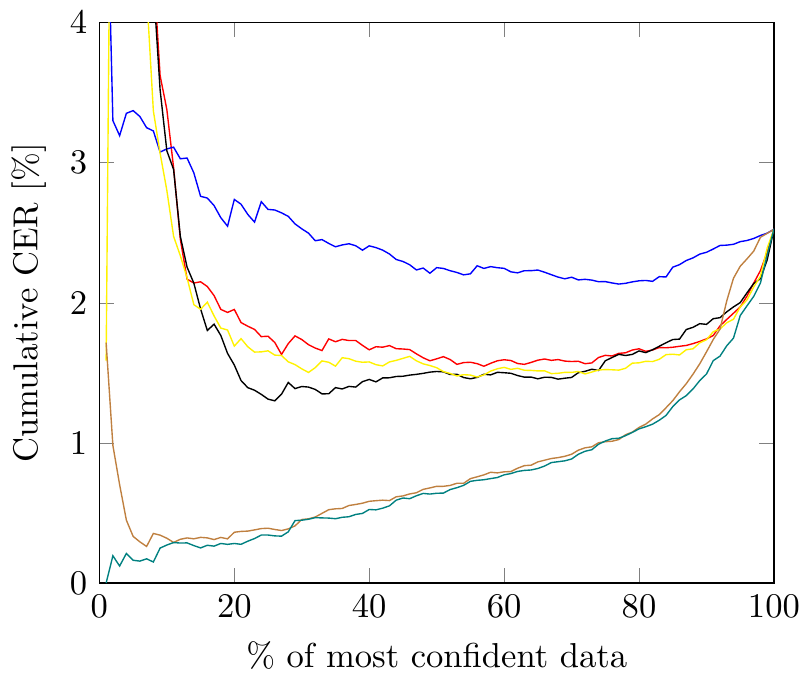}
    }
    \\
    \subcaptionbox{%
        Printed, small\label{fig:conf-printed-small}}
        {\includegraphics[width=0.45\linewidth]{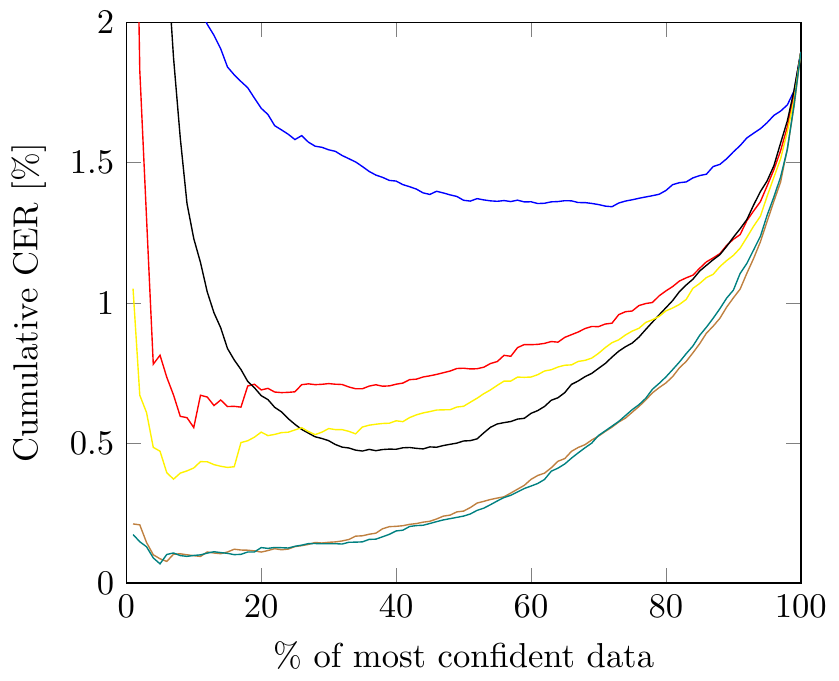}
    }
    \qquad
    \subcaptionbox{%
        Printed, big\label{fig:conf-printed-big}}
        {\includegraphics[width=0.45\linewidth]{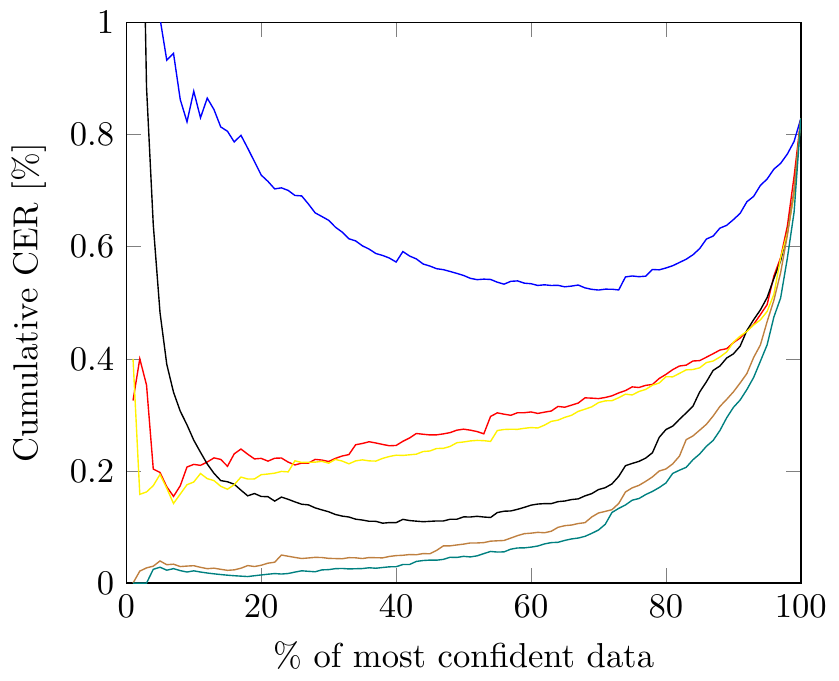}
    }
    \\[1em]
    \includegraphics[width=0.75\linewidth]{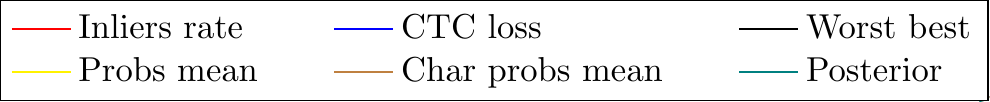}
    \caption{%
        CER as a function of the size of considered confident data.
        Note the different scale of the error rate.
    }
    \label{fig:metrics}
\end{figure}

\begin{table}[p]
    \centering
    \caption{%
        AUCs for individual confidence measures.
    }\label{tab:aucs}
    \begin{tabular}
        {
        @{\extracolsep{12pt}}@{\kern\tabcolsep}
        lcccc}
        \toprule
        \multirow{2}{*}{Confidence measure} & \multicolumn{2}{c}{Handwritten} & \multicolumn{2}{c}{Printed} \\
                        \cline{2-3}
                        \cline{4-5}
                        \addlinespace[0.1cm]
                    & Small    & Big          & Small    & Big  \\
        \midrule
        CTC loss            & 7.082 & 2.453 & 1.557 & 0.656 \\
        Posterior           & 2.798 & \textbf{0.754} & \textbf{0.429} & \textbf{0.109} \\
        Probs mean          & 4.218 & 2.035 & 0.747 & 0.283 \\
        Char probs mean     & \textbf{2.657} & 0.844 & 0.431 & 0.131 \\
        Inliers Rate        & 4.053 & 2.277 & 0.896 & 0.305 \\
        Worst best          & 5.443 & 2.428 & 1.189 & 0.260 \\
        \bottomrule
    \end{tabular}
\end{table}

Having the seed optical models, we use them to evaluate the confidence measures proposed in Section~\ref{sec:confidences-def}.
We do so by sorting the validation lines by their confidence scores and then calculating the CER on the most confident subsets of increasing size.
The resulting progress of the CER is visualized in Figure~\ref{fig:metrics}.
To assess the quality of the measures quantitatively, we calculate the area-under-curve (AUC) for each confidence measure.
We report the AUCs in Table~\ref{tab:aucs}.
The \emph{Transcription posterior probability} comes out as the best measure, having the best overall AUC as well as very consistent behaviour for all operating points.
Therefore, we picked it to perform the data selection in the self-training experiments.

\begin{figure}[t]
    \centering
    \subcaptionbox{%
        Handwritten\label{fig:est-cer-bentham}}
        {\includegraphics[width=0.48\linewidth]{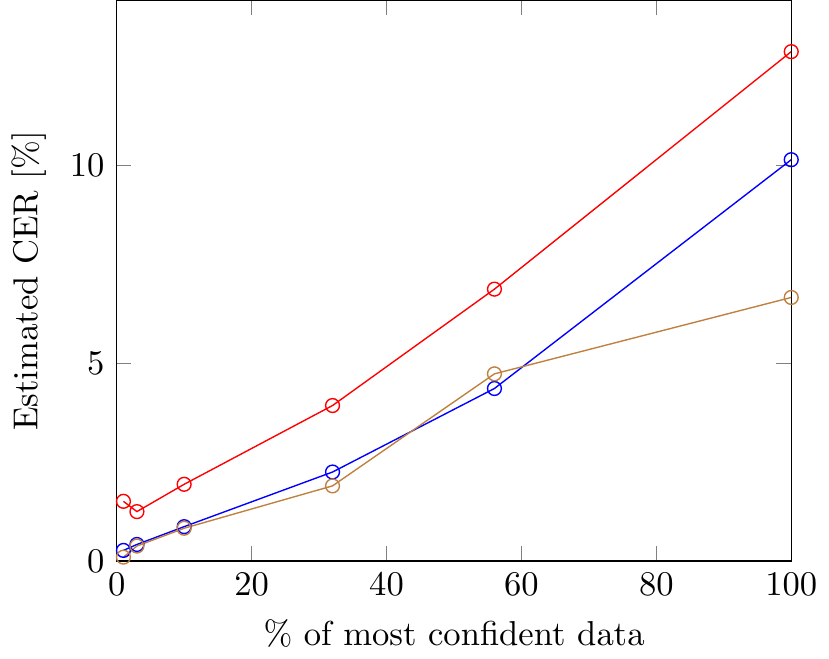}
    }
    \subcaptionbox{%
        Printed\label{fig:est-cer-printed}}
        {\includegraphics[width=0.48\linewidth]{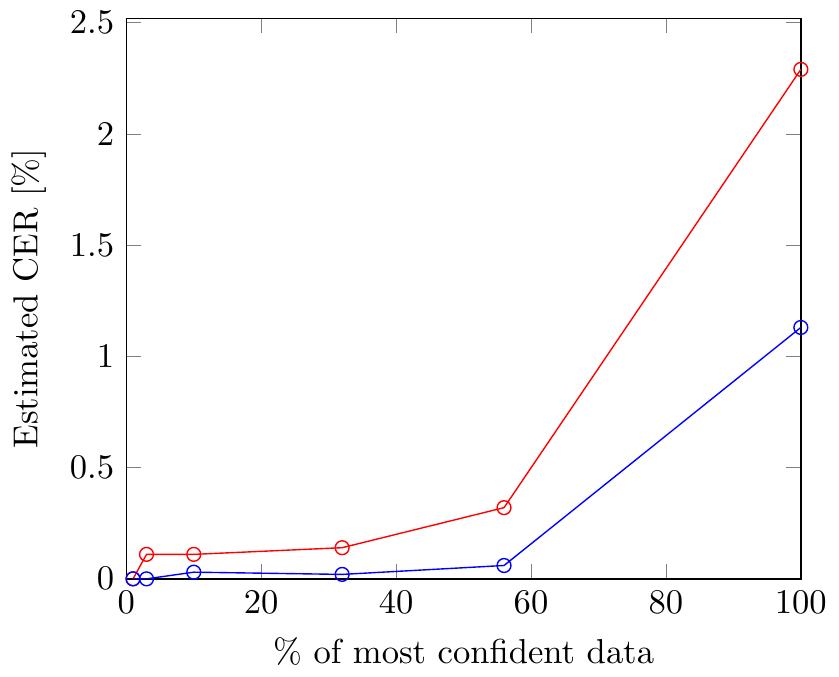}
    }
    \\[1em]
    \includegraphics[width=0.70\linewidth]{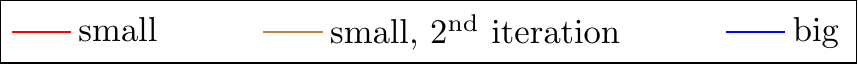}
    \caption{
        CER of portions of unannotated data, as estimated by the confidence measure.
    }\label{fig:est-cers}
\end{figure}

Additionally, we use it to estimate the CER of the individual portions of the MA data.
To estimate the CER of any text line, we find its 10 nearest neighbours in the validation set by the confidence score, and average their CERs.
The estimates are summarized in Figure~\ref{fig:est-cers}.
According to the confidences, the data varies from easy to rather difficult, esp. in case of the handwritten dataset.
The slightly inconsistent output of 2\textsuperscript{nd} iteration is probably a consequence of confirmation bias, as model has already been trained on parts of this data.

\subsection{Effect of introducing machine-annotated data}\label{sec:ss-exp}

The core of our contribution is in demonstrating the improvements from adapting the text recognition system to the unannotated part of the target domain.
We report how much the OCR improves when additionally presented 1, 3, 10, 32, 56, and 100\,\% of the most confident MA data.
We perform these experiments in two separate branches.

\begin{figure}[t!]
    \centering
    \subcaptionbox{%
        Handwritten, Optical model\label{fig:ss-cer-hwr-ocr}}
        {\includegraphics[width=0.48\linewidth]{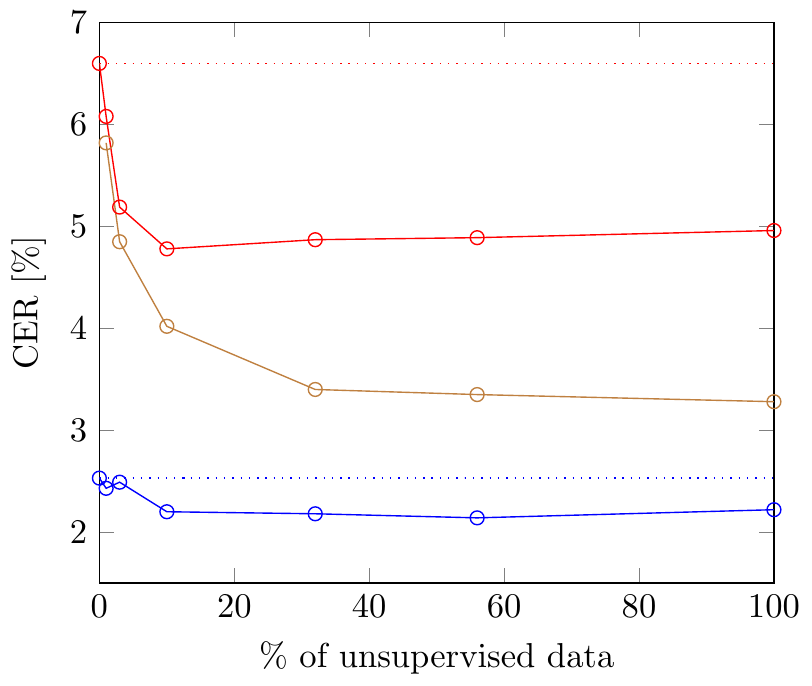}
    }
    \subcaptionbox{%
        Handwritten, Langauge model\label{fig:ss-cer-hwr-lm}}
        {\includegraphics[width=0.48\linewidth]{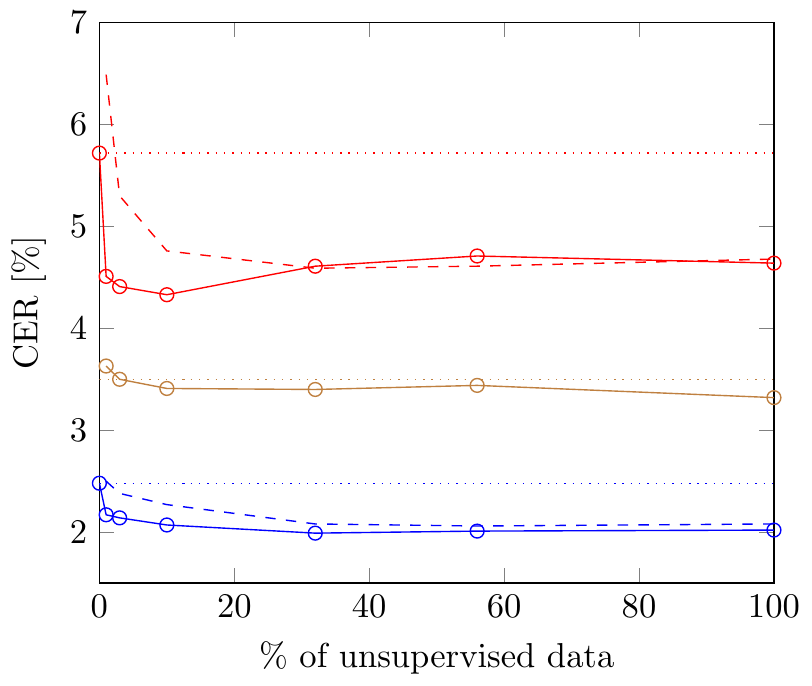}
    }
    \\[1em]
    \subcaptionbox{%
        Printed, Optical model\label{fig:ss-cer-printed-ocr}}
        {\includegraphics[width=0.48\linewidth]{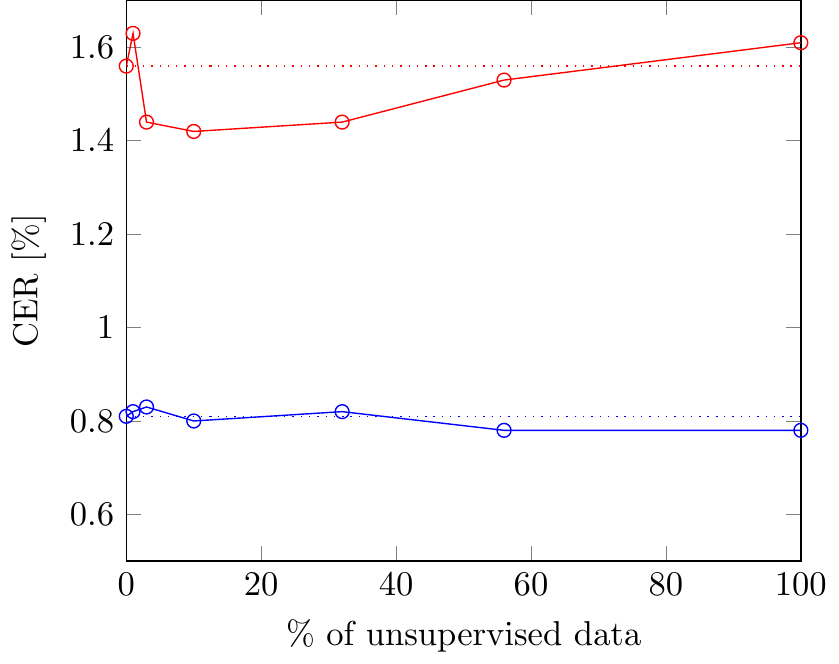}
    }
    \subcaptionbox{%
        Printed, Language model\label{fig:ss-cer-printed-lm}}
        {\includegraphics[width=0.48\linewidth]{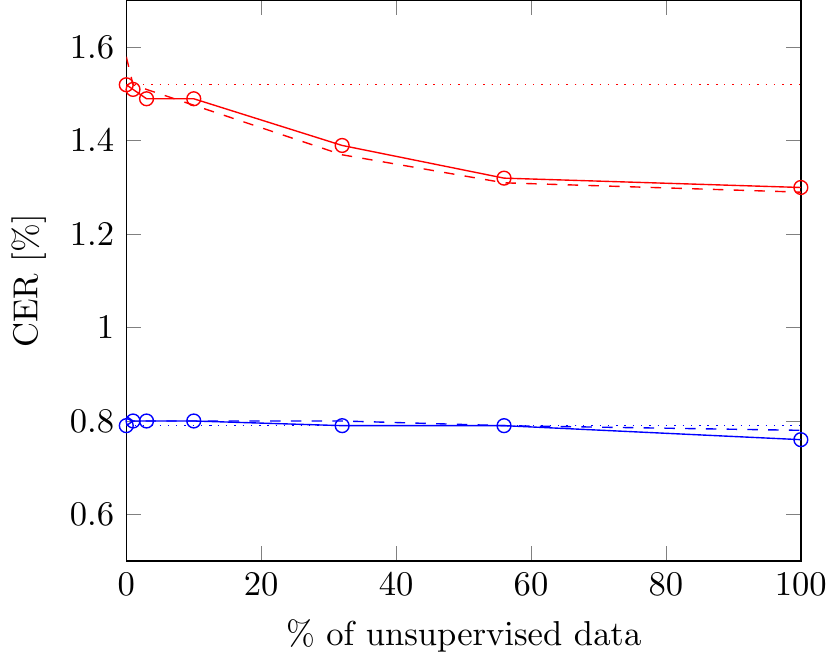}
    }
    \\[1em]
    \includegraphics[width=0.70\linewidth]{figures/cer-legend.pdf}
    \caption{
        System performance in different experimental branches and datasets.
        Dotted line shows the corresponding baseline.
        Dashed line in the LM branch denotes decoding with LM trained on target domain only.
    }\label{fig:ss-cers}
\end{figure}

The first branch focuses on the optical model (OM) only, where output is obtained using greedy decoding.
In this case, the MA data is mixed with the seed data and a new OM is trained from scratch.
Results are shown in Figures~\ref{fig:ss-cer-hwr-ocr} and~\ref{fig:ss-cer-printed-ocr}.

The second branch keeps the seed OM fixed and studies the effect of adding the MA data to the language model.
Training LMs is done in stages:
(1) The LM is pretrained on the related domain data, (2) it is further trained on the given amount of MA data, and (3) it is fine-tuned to the target domain annotations.
The LM is introduced into prefix search decoding as shown in Eq.~\eqref{eq:decoder}.
Results are shown in Figures~\ref{fig:ss-cer-hwr-lm} and~\ref{fig:ss-cer-printed-lm}.

Additionally, we demonstrate the possibility to obtain further gains from a second iteration on the handwritten small setup, which is the most challenging one.
In this case, we take the fusion of the best performing OM with best performing LM as a new seed system to transcribe the unannotated data.
This setup is referred to as \emph{small, 2\textsuperscript{nd} iteration}.

From the validation results in Figure~\ref{fig:ss-cers}, we see that, with the exception of the printed big setup, adding the MA data provides significant improvement to the performance of the system and this improvement is smooth with respect to the amount of MA data.
We assume that in the printed big setup, we are already hitting the level of errors in the annotations.

When adapting the OMs, it is more beneficial to only take a smaller portion of the MA data in the small setups.
On the other hand, in the big setups and the second iteration of the handwritten small setup, the OM improves more on the larger portion of the MA data.

When adapting the LMs, largest improvement is achieved by introducing all of the MA data in all cases but one:
It is optimal to only take 10\,\% of the MA data in the handwritten small setup (Fig.~\ref{fig:ss-cer-hwr-lm}).
The reason, why this setup is special, is the combination of insufficient amount of target domain annotations and the comparatively low quality of the seed system\,--\,when pretraining data is not available or when the seed system is more accurate, it is again the most beneficial to consider all of the available MA data.
In contrast to the handwritten setups, related domain data brings no significant difference in the printed setups.
We expect this to be a consequence of relatively large amount of available target domain annotations and the good accuracy of the machine annotations.

Comparing LM and OM, we hypothesize the difference in behaviour in the small setups comes from the fact that while LM is a generative model which is expected to produce a less sharp distribution in the face of noisy data, the discriminative OM is prone to learning a wrong input-output mapping.
This is in line with the observation that the OM benefits much more from the second iteration, where the error rate of the machine annotations is lower.
Additionally, it probably benefits from the different knowledge brought in by the LM, which may break the limit of confirmation bias.
On the other hand, the LM is simply exposed to less noisy data, but not to a new source of knowledge.

Finally, we take the best performing amounts of MA data and compare all the relevant combinations of models on the test data, as summarized in Table~\ref{tab:final-cers}.
In line with the findings on the validation set, the fusion of the best OM with the best LM performs the best in all setups, providing massive gains over the baseline.

\begin{table}[t]
    \centering
    \caption{%
        Final comparison of results on test datasets, reported as CER [\%].
        Optimal size of MA data as well as other hyperparameters are selected as per validation performance.
    }\label{tab:final-cers}
    \begin{tabular}{
        @{\extracolsep{12pt}}@{\kern\tabcolsep}
        llccccc}
        \toprule
         &    & \multicolumn{3}{c}{Handwritten} & \multicolumn{2}{c}{Printed} \\
                        \cline{3-5}
                        \cline{6-7}
                        \addlinespace[0.1cm]
        OM             & LM            & Small     & Small-2   & Big       & Small & Big  \\
        \midrule
        seed            & none          & 6.43      & 4.41      & 2.58      & 1.03  & 0.29  \\
        optimal         & none          & 4.41      & 2.92      & 2.07      & 0.78  & \textbf{0.27}  \\
        \midrule
        seed            & seed          & 5.34      & 3.27      & 2.53      & 0.99  & \textbf{0.27}  \\
        seed            & optimal       & 4.17      & 3.21      & 2.02      & 0.76  & \textbf{0.27}  \\
        optimal         & optimal       & \textbf{3.27}      & \textbf{2.88}      & \textbf{1.94}      & \textbf{0.64}  & \textbf{0.27} \\
        \bottomrule
    \end{tabular}
\end{table}

\subsection{Masking augmentation effect}\label{sec:masking-exp}
Besides the experiments with unannotated data, we also explored the effect of the masking augmentation.
The aim of the first experiment is to compare the effect of the masking augmentation with the traditional augmentations (affine transformations, change of colors, geometric distortions,  etc.)
We conducted this experiment on the related domain only, results are in Table~\ref{tab:augmentations_comparison}.
The results show that the masking does improve the recognition accuracy, however the impact is smaller than that of a combination of traditional approaches and is not significant in the case of the easy printed data.

\begin{table}[t]
    \centering
    \caption{%
        Augmentations comparison in CER [\%]
    }\label{tab:augmentations_comparison}
    \begin{tabular}{
        @{\extracolsep{12pt}}@{\kern\tabcolsep}
        lcc}
        \toprule
        Augmentation    &  READ             & IMPACT \\
        \midrule
        None            &  5.28             &  0.66  \\
        Masking         &  3.91             &  0.65  \\
        Traditional     &  3.69             &  \textbf{0.58}  \\
        Both            &  \textbf{3.20}    &  \textbf{0.58}  \\
        \bottomrule
    \end{tabular}
\end{table}

The second experiment aims to determine the effect of distribution of the masking across the text line.
We did so by training three models with different settings of the success probability and the maximal width of the masked region.
In order to isolate the effect of frequency of masking, we designed the settings so that the expected volume of the masking remains the same.
This experiment was performed on the combination of the related domain, the annotated target domain, and 32 \% of MA data in the first iteration of our self-training pipeline.
The settings and the results are presented in Table~\ref{tab:masking_types}.
Overall, the impact of the masking augmentation is very robust to the frequency of the masking.

\begin{table}[]
    \centering
    \caption{%
        Results of different masking augmentation settings in CER [\%].
        The success probability and the width of the region are denoted as $p$ and $W$ respectively.
        Base-probability refers to the setting used in self-training experiments.
    }\label{tab:masking_types}
    \begin{tabular}{
        @{\extracolsep{12pt}}@{\kern\tabcolsep}
        lcccccc}
        \toprule
        & \multicolumn{2}{c}{Settings} & \multicolumn{2}{c}{Handwritten} & \multicolumn{2}{c}{Printed} \\
                        \cline{2-3}
                        \cline{4-5}
                        \cline{6-7}
                        \addlinespace[0.1cm]
                        & $p$ & $W$ & Small  &  Big   & Small  & Big  \\
        \midrule
        Without masking     & 0                     &  ---  &  5.30  &  2.27  &  1.51  &  0.86  \\
        Half-probability    & \multicolumn{1}{r}{$2.5 \times 10^{-3}$} & $[5, 80]$  &  4.98  &  \textbf{2.13}  &  1.49  &  0.85  \\
        Base-probability    &  \multicolumn{1}{r}{$5 \times 10^{-3}$}  &  $[5, 40]$  &  \textbf{4.87}  &  2.18  &  \textbf{1.44}  &  \textbf{0.82}  \\
        Double-probability  & \multicolumn{1}{r}{$10 \times 10^{-3}$}  &  $[5, 20]$  &  4.89  &  2.15  &  1.50  &  0.83  \\
        \bottomrule
    \end{tabular}
\end{table}

\section{Conclusion}
In this paper, we have described and experimentally validated a self-training adaptation strategy for OCR.
We carefully examined several confidence measures for data selection.
Also, we experimented with image masking as an augmentation scheme. 

Our self-training approach led to reduction in character error rate from 6.43\,\% to 2.88\,\% and 1.03\,\% to 0.64\,\% on handwritten and printed data respectively.
As the confidence measure, the well-motivated transcription posterior probability was identified as the best performing one.
Finally, we showed that the proposed masking augmentation improves the recognition accuracy by up to 10\,\%.

Our results open way to efficient utilization of large-scale unlabelled target domain data.
This allows to reduce the amount of manual transcription needed while keeping the accuracy of the final system high.
As a straight-forward extension of this work, our self-training method can be applied to sequence-to-sequnce models.
Regarding the masking augmentation, it would be interesting to specifically assess its impact on learning from the machine annotated data and to test whether it truly pushes the CTC optical model towards learning a better language model.
Finally, data selection method considering also novelty of the line could lead to better utilization of the unannotated target domain data data.
%
%
%
\bibliographystyle{splncs04}
\bibliography{mybib}

\end{document}